# Reinforcement Learning for Accelerated Aerodynamic Shape Optimisation

**F Sobieczky[1], A Lopez[1], E Dudkin[1], C Lackner[3], M Hochsteger[3], B Scheichl[2], H Sobieczky[2]**

[1] Software Competence Center Hagenberg (SCCH), Softwarepark 32a, 4232 Hagenberg i. Mkr., AT
[2] Institut für Strömungsmechanik und Wärmeübertragung, TU Wien, Getreidemarkt 9, 1060 Vienna, AT
[3] CERBSim GmbH. Taubstummengasse 11. 1040 Vienna, AT

**Keywords:** Reinforcement learning, aerodynamic shape optimization, computational domain reduction

**Abstract:** We introduce a reinforcement learning (RL) based adaptive optimization algorithm for aerodynamic shape optimization focused on dimensionality reduction. The form in which RL is applied here is that of a surrogate-based, actor-critic policy evaluation MCMC approach allowing for temporal 'freezing' of some of the parameters to be optimized. The goals are to minimize computational effort, and to use the observed optimization results for interpretation of the discovered extrema in terms of their role in achieving the desired flow-field. By a sequence of local optimized parameter changes around intermediate CFD simulations acting as ground truth, it is possible to speed up the global optimization if (i) the local neighbourhoods of the parameters in which the changed parameters must reside are sufficiently large to compete with the grid-sized steps and its large number of simulations, and (ii) the estimates of the rewards and costs on these neighbourhoods necessary for a good step-wise parameter adaption are sufficiently accurate. We give an example of a simple fluid-dynamical problem on which the method allows interpretation in the sense of a feature importance scoring.

## 1. *Introduction*

**Related work** Reinforcement learning (RL, [16, 17]) is a method from dynamic programming and has evolved into a versatile complex of different methods, each being able to employ the advantages of the ability to learn a strategy of how to solve a similar optimization problem. In continuous optimization, such as aerodynamic shape optimization (ASO), RL has recently gained attention due to its capabilities of handling changing conditions [18]. The need to adapt to changing conditions during an optimization process is a standard challenge in optimization, and learning a strategy to cope with these changes is a key feature of all dynamic programming methods. This property of all agent-based Markov decision processes helps defining rules to cope with changing environments during the optimization. This is particularly useful in fluid mechanics where the predictive power of deep learning models is used to approximate RF's value functions [19]. In the present context, we will use a simple analytical surrogate with statistical parameter estimation to perform the value function approximation. It will be shown in this paper that ASO involving computationally expensive simulations can be accelerated by using RL, if there is *some* model for local predictions of the flow-field changes associated with fluid-dynamical parameter changes. As our focus is to accelerate the optimization process, our method belongs to the field of model order reductions (MOR, or ROM). Like CWROM (component-wise ROM, [20]), we evaluate each component, individually, as to whether it needs further update. Similar to the approach chosen in TRPO (Trust region policy optimization, [21]), our method estimates the changes of the solution on the previous performance on bounded subregions of the parameter space, acting as the RL state space.

**Context**: In Computational Fluid Dynamics (CFD), often spatially complicated and temporally periodically forced flows, as emerging in engineering real-life tasks, are encountered. Such flows involve very large characteristic Reynolds numbers and are thus typically associated with developed turbulence. In order to restore at least the important imposed large-scale unsteadiness, one approaches them either by the unsteady extension of classical RANS methods, i.e. URANS [1f, 1e], or by filtering methods restrict the modelling of turbulence to its fine structure, i.e. Large-Eddy Simulation (LES) using so-called sub-grid models, or, in still rarer cases, by Direct Numerical simulation (DNS) that attempts fully deterministic solutions of the Navier–Stokes equations. As the latter is still restricted to rather simple geometries and LES involves subtle intricacies when separating the physical (modelling) error from the methodical (numerical) one, (U)RANS provides still the wide-spread means of choice for simulating flows in industrial applications. For both the RANS and the LES approaches, however, machine learning methods have been adopted to increase the performance and/or aid in gaining a deeper physical understanding of the predictions under parameter variations [22, 23, 24]. Specifically, reinforcement learning has been successfully used for solving dynamic optimization problems [10, 11, 12, 13, 14]. Hybrid approaches combining computational techniques with physical modelling [8] may prove beneficial for overcoming the problems associated with high dimensionality of parameter spaces in optimization problems of physical surrogate models [9]. Reducing the computational resolution of CFD optimization based on sensitivity results has been studied in [15]. We propose a different method to tackle this task efficiently by employing a model-based Monte-Carlo approximation of the value function [16].

**General problem description**: The problem to be solved in this work is the shape optimization of a flow rectifier changing an incoming inhomogeneous multidirectional flow into a homogeneous unidirectional flow (parallel to the X-axis). It arises in the context of the subsonic (and first treated as incompressible) airflow of an air source heat pump upstream of the heat exchanger. Specifically, too pronounced inhomogeneities in terms of the turbulence level intensities or simply a lacking collinearity with the mean-flow direction are known to severely compromise the efficiency of the heat transfer [1]. In general, a homogeneous, steady flow perpendicular to the inflow plane of the heat exchanger is desired. Achieving this poses a formidable, (at least) threefold challenge of non-linear large-scale optimization: (i) establishing a proper comprehensive flow model (and a rationally devised reduced-order surrogate model); (ii) its treatment by full CFD (and semi-analytical one of the reduced one); (iii) setting up an efficient and interpretable optimization strategy based on RL.

**Preparatory model**: To keep the use case as concise and illustrative as possible, we consider a model problem of drastically reduced complexity that yet allows for demonstrating the applicability of our optimization strategy: when adopting Cartesian coordinates and X and Z, the planar, isoviscous Stokes flow past an airfoil placed fully in a straight channel of finite length aligned with the X-direction with pervious walls, where periodic boundary conditions are met. The thereby formed array of profiles might be interpreted as an unwound sequence of directing blades of an axial rotor in an artificial low-Reynolds-number flow. The inflow is uniform but also exhibits a component in the wall-normal or Z-direction. Natural or dynamic outflow conditions state vanishing total stress, which arise from the weak formulation of the problem. This is most suitable for its numerical solution by a finite-element discretisation, using the open-source software environment *Netgen/NGSolve* [5]. It is our objective to redirect the flow such that (i) along a line parallel to the Z-axis behind the airfoil (dashed line), there is as little Z-directional flow components as possible $\boldsymbol{L^2}$-norm, and (ii) the magnitude of the flow on that cross-section is as evenly distributed as possible (see Figure 1, Left).

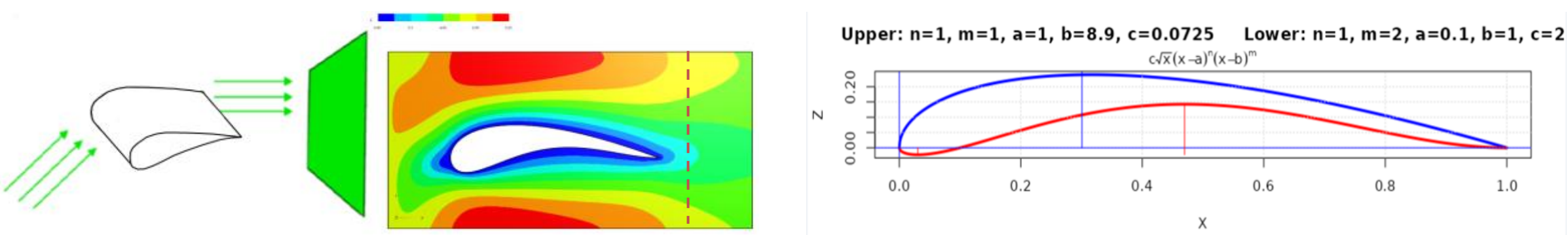

*Figure 1: Left/Middle: Sketch and solution [5] of a two-dimensional flow problem for a blade-induced stratification of a flow field with positively inclined incoming direction. Right: Reduced PARSEC parametrisation of a profile for the redirecting blade. The two optimization parameters are b(here: 8.9) from the upper (blue), and a (0.1) from the lower side (red) of the profile. By an analytical condition on the thickness of the profile, the leading coefficient c is determined, implicitly.*

The paper is organized in the following way: Section 2 contains the definition and theoretical background of the reinforcement learning method used, while in Section 3 it is applied to the use case described in the first section. Section 4 contains discussion, conclusion and outlook in front of the obtained results.

## 2. *Method*

**PARSEC**. In the context of an air source heat pump efficiency problem [1], the homogenization of the outward flow-field of a channel is pursued by placing guiding blades into the airflow to stratify the air before it reaches the heat exchanger. We follow the approach of continuous interpretable aerodynamic shape optimization (ASO) with Reinforcement Learning (RL) with a geometry model $\boldsymbol{M}: \boldsymbol{\Theta} \rightarrow \boldsymbol{\Theta}$, mapping a given state of geometric shape parameters into the new, changed state: $\boldsymbol{M}(\boldsymbol{\theta_n}) = \boldsymbol{\theta_{n+1}}$. The state description includes a flag for each parameter to be either changeable or fixed during the subsequent adaption. To decide about the specific state of this flag more easily, an *interpretable* geometry parametrization for airfoil design parameters, called PARSEC [2], is employed. The strategy to be learned by RL is to freeze some parameters from the PARSEC and camber line dimensions [6] to fewer degrees of freedom to increase the performance to a feasible calculation speed of the full configuration's optimization. With PARSEC, the upper and lower contour of the airfoil is modeled by

$$\boldsymbol{Z(x) = \sqrt{x} \cdot p(x) = \sqrt{x} \cdot \sum_{i=0}^{5} a_i x^i}$$

a product of a degree 5 polynomial with the square root function. Note that the four digit NACA family is the *sum* of the square root with a degree 5 polynomial.

A suitable representation of the PARSEC parameters is chosen by Reduced PARSEC, defined in Figure 1 (Right). The algorithm to reduce the dimensionality of the changeable PARSEC parameter space is indicated in Figure 2 (Left). Figure 2 (Right) shows the original PARSEC design parameters and their relationship with Reduced PARSEC, in which the associated polynomial is factorized into its linear factors. Freezing the parameters in this representation corresponds to allowing only a subset of the roots to be varied.

**Reduced PARSEC (with fixed roots)**

**Idea**: In PARSEC parametrization use root-factorization representation of polynomial and **fix roots**. Get Interpretation from change in PARSEC-Prmts.

**Recipe:** (During one cycle)

1. Choose PARSEC-Parameter $\theta \in \Theta$.
2. Calculate Coefficients $\boldsymbol{a^i}, i \in \{1 \dots 6\}$.
3. Find roots $x_i$ of $p(x)$.
4. Observe **stability** of roots.
5. Choose action: If root stable, reduce parameter space by 'fixing' this root. Results in smaller # of dimensions.
6. Calculate resulting parameters $\theta' \in \Theta$.
7. **Interpret** parameter changes $\Delta\theta = \theta' - \theta$.

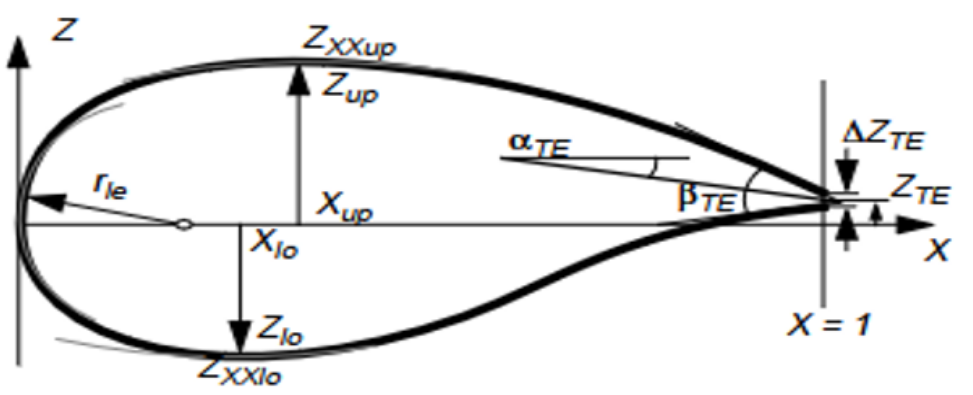

$$z_{up} = \sum_{i=1}^{n=6} a_{up}^i \cdot x^{i-\frac{1}{2}}, \qquad z_{lo} = \sum_{i=1}^{n=6} a_{lo}^i \cdot x^{i-\frac{1}{2}}$$

**Reduced PARSEC:**

$$Z(x) = \sqrt{x}p(x) = \sqrt{x}c\Pi_{i\in\{1\dots5\}}(x - x_i)$$

Roots $x_i$ plus amplitude $c$ are an equivalent representation. It is useful for *reducing* the number of degrees of freedom.

*Figure 2 – Left: Policy re-evaluation after each cycle by using notion of 'stable parameter': If a parameter is considered stable it is removed from the set of changeable parameters. Right: Definition of PARSEC [2, 3] and Reduced PARSEC. Reduced PARSEC used the polynomial $p(x)$'s factorization into its linear factors.*

In the present note, we use a simplification of this Reduced PARSEC-parametrization by considering a family of airfoils also generated by the product of the square root function with a polynomial, however one of lesser degree. The airfoil shown in Figure 1 is defined by an upper and lower Reduced PARSEC family with respective polynomials $\boldsymbol{p(x)}$ of degree two and three.

**Reinforcement Learning**. In reinforcement learning there is a state space $\boldsymbol{S}$, a set of actions $\boldsymbol{A}$, as well as a family of conditional probability measures $\boldsymbol{\pi}: \boldsymbol{S} \times \boldsymbol{A} \rightarrow [\boldsymbol{0}, \boldsymbol{1}]$, the policies, evaluating by $\boldsymbol{\pi}(\boldsymbol{a_n}|\boldsymbol{s_n})$ a potential step using action $\boldsymbol{a_n} \in \boldsymbol{A}$ from state $\boldsymbol{s_n}$ to $\boldsymbol{s_{n+1}}$. The decision is based on the evaluation of an instantaneous reward $\boldsymbol{R}(\boldsymbol{s_{n+1}}, \boldsymbol{a_n}, \boldsymbol{s_n})$, with the reward function $\boldsymbol{R}: \boldsymbol{S} \times \boldsymbol{A} \times \boldsymbol{S}$. We will use the reward in the sense of a penalty (negative reward), meaning that our optimum occurs for small values, and in the form of a function depending only on the target state, $\boldsymbol{s_{n+1}}$. The action leading to it is automatically $a_n = s_{n+1} - s_n$ and must reside in the currently valid action set $\boldsymbol{A}$.

The complete evaluation of a single step from one state to the next involves considering the accumulated and discounted reward, $\sum_{t=0}^{\infty} \gamma^{t} R(s_{n+t})$, that is, the random variable of the currently estimated value of the future rewards. The state space used here is a subset of the reasonable choices of PARSEC parameter vectors, by which we mean the cartesian product of finite sets of suitable choices of parameter values, typically a finite grid of vectors with coordinates from sets of the form $\left[\boldsymbol{\theta_{i_{min}}}, \boldsymbol{\theta_{i_{max}}}\right] \cap \boldsymbol{\delta_i Z}$ with an appropriate interval size between neighbouring values of the $\boldsymbol{i}$-th parameter. *It is our goal to accelerate the optimization process by allowing actions to be chosen only by such nearest neighbours of this grid if the change of this specific parameter also contributes to a significant change in the reward.*

We use a Monte-Carlo method to make the optimal choice of the parameter update. The proposed RL framework is modeled as a Markov Decision Process MDP, represented by a random walk on the state space When a parameter update occurs, it starts from the knowledge of the reward at a given parameter $\boldsymbol{\theta_n}$. All parameters in $N(\theta_n)$ are used as the initial state in a cycle of the MDP. There is no policy change during a cycle, but where the same action is used for a specific number of steps. Along this notion, we choose two scales, one *microscopic scale* referring to the neighbourhood of a state in terms of the chosen computational grid, and a *mesoscopic scale* referring to the larger neighbourhood $\boldsymbol{N(\theta_n)}$ of a given state $\boldsymbol{\theta_n} \in \boldsymbol{S}$ within which the cycle with the same policy runs until the new state $\boldsymbol{\theta_{n+1}} = \boldsymbol{M(\theta_n)}$ is chosen. When this adaptation takes place, also a policy evaluation is carried out for the next cycle, in which $\boldsymbol{\theta_{n+1}}$ is the new initial state. An estimation of the best policy is carried out based on aggregated information on the different possible policies in the previous (n-th) cycle.

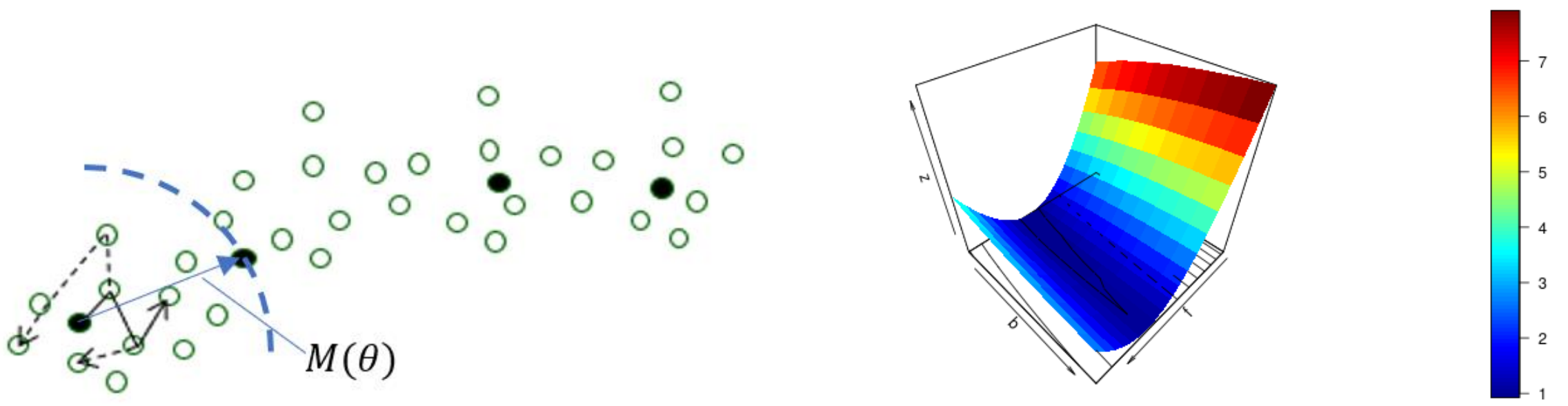

*Figure 3 – Left: Sketch of the microscopic and mesoscopic scale: The blue dashed line refers to $N(\theta_n)$ on which a single cycle with the same chosen policy takes place. The optimal state $\theta_{n+1}^*$ is chosen as the initial state for the subsequent cycle. Right: Calculation of the exact reward function $z = R(b, f)$ where $b$ is the larger of two real valued roots of the polynomial $p(x)$ in the Reduced PARSEC representation of the airfoil. $f$ is the larger root of the camber function $CAM(x) = 0.3x(f - x)$.*

The set $\boldsymbol{N(\theta_n)}$ acts therefore as the state space of the Markov Decision Process (MDP) with initial state $\boldsymbol{\theta_n}$ measured on the mesoscopic time $\boldsymbol{n} \in \boldsymbol{N}$. The schematics of this two-scale sequence of RL-methods is indicated in Figure 3 (Left), which shows the subsequence of states for which an actual CFD simulation is carried out by black dots. These also act as initial states for the subsequent cycle of the MDP. The larger steps possible on the mesoscopic scale are characteristic of the reduced computational effort in CFD simulations. If the neighbourhood $\boldsymbol{N(\theta)}, \boldsymbol{\theta} \in \boldsymbol{S}$ is chosen too large, it is likely that the lack of policy re-evaluation misleads the agent and prohibits finding the global extremum on the mesoscopic scale.

The allowed actions involve changes of parameters among its neighbours on a initially chosen cartesian grid of discrete steps. The MDP thus becomes a random walk on this grid. The initial state $\boldsymbol{\theta_n}$ changes after $\boldsymbol{t}$ discrete steps into $\boldsymbol{\theta_n} + \boldsymbol{\Delta\theta_t}$ with $\boldsymbol{\Delta\theta_t} = \sum_{j=1}^{t} \boldsymbol{X_j}$, with increments $\boldsymbol{X_j}$ chosen from the possible actions in the currently chosen set of actions $\boldsymbol{A_i}$ with probability given by the current policy $\boldsymbol{\pi_i}$. Re-evaluation of the policy only between two consecutive cycles makes the random walk have the same transition probability $\boldsymbol{P_i}$ (with $(\boldsymbol{P_i})_{\boldsymbol{\theta},\boldsymbol{\theta'}} = \boldsymbol{\pi_i}(\boldsymbol{a} = \boldsymbol{\theta'} - \boldsymbol{\theta} | \boldsymbol{s} = \boldsymbol{\theta})$ during a single cycle.

The reward function for the current state $\boldsymbol{\theta} = \boldsymbol{\theta_n} + \boldsymbol{\Delta\theta_t}$ is, for the sake of a homogeneous flow-field parallel to the X-axis, chosen to be a negative cost, $\boldsymbol{R_t(\theta_n)} = \boldsymbol{R_1(\theta_n + \Delta\theta_t)} + \boldsymbol{R_2(\theta_n + \Delta\theta_t)}$ **is** given by

$$\boldsymbol{R_1(\theta_n + \Delta\theta_t)} = \frac{1}{|E|} \sum_{x \in E} \frac{|\boldsymbol{u_2}(x, \boldsymbol{\theta} + \boldsymbol{\Delta\theta_t})|^2}{(|\boldsymbol{u_1}(x, \boldsymbol{\theta} + \boldsymbol{\Delta\theta_t})|^2 + |\boldsymbol{u_2}(x, \boldsymbol{\theta} + \boldsymbol{\Delta\theta_t})|^2} \tag{1}$$

and

$$\boldsymbol{R}_{2(\boldsymbol{\theta_n} + \boldsymbol{\Delta\theta_t})} = \max_{x \in E} |\boldsymbol{u}(\boldsymbol{x}, \boldsymbol{\theta_n} + \boldsymbol{\Delta\theta_t})| - \min_{x \in E} |\boldsymbol{u}(\boldsymbol{x}, \boldsymbol{\theta_n} + \boldsymbol{\Delta\theta_t})| \,. \tag{2}$$

Here, $\boldsymbol{R_1}$ is the mean ratio of the squared vertical component to the magnitude squared of the velocity vector on the subset $\boldsymbol{E}$ of the air channel representing the flow field behind the blade, and $\boldsymbol{R_2}$ the spread on $E$. The 'plane' $\boldsymbol{E}$ is given by a vertical line in our 2-dimensional setting, as shown in Figure 1 (Left). Figure 3 (Right) shows a plot of $\boldsymbol{R(\theta)} = \boldsymbol{R_1(\theta)} + \boldsymbol{R_2(\theta)}$ for a special case of other parameters than $b$, and $f$. The difference in prominence of the two parameters is seen: Even though there is a distinct minimum, it is approached from the $f$-side by a much stronger gradient than from the $b$-side.

Instead of allowing changes between all nearest neighbours in the parameter grid $\boldsymbol{S}$, we define $\boldsymbol{A}$ to be the set of finite functions

$$\boldsymbol{A} = \left\{ \eta : \{1, \ldots, d\} \to \times_{\boldsymbol{i=1}}^{\boldsymbol{d}} \{-\delta_i, 0, \delta_i\} \right\},$$

where $\boldsymbol{d}$ is the maximal number of changeable parameters (i.e., the dimension of the parameter grid). So, $\boldsymbol{A}$ is identical to the set of subsets of $\{\boldsymbol{1}, \ldots \boldsymbol{d}\}$, which include the indices of the parameter-vector, which are allowed to change to its nearest neighbour (at distance $\delta_i$ away, in either direction). Changing them means adding $\delta_i$ or $-\delta_i$ to them. For example, if $\boldsymbol{d} = \boldsymbol{2}$, the case we will consider in the use case, there is $\boldsymbol{A_1} = \{-f, +f\}$, $\boldsymbol{A_2} = \{-b, b\}$, $\boldsymbol{A_3} = \{-f, +f, -b, +b\}$. The set of allowed actions thus corresponds to 'back' ($-\delta_i$) and 'forth' ($+\delta_i$) on the power set of the set of dimensions.

If $\boldsymbol{A_i}$ is the set of current possible actions (allowing only a specific subset of parameters to be changed) and $\boldsymbol{\pi_i}$ the corresponding policy, the method of finding a new parameter $\boldsymbol{\theta_{n+1}}$ involves minimizing the current expected discounted return, given by

$$\boldsymbol{V}_{\boldsymbol{\pi_i}}(s) = \boldsymbol{E}_{\boldsymbol{\pi_i}}[\textstyle\sum_{\boldsymbol{t=0}}^{\infty} \boldsymbol{\gamma^t R_{t+1}} \,| s] = \sum_{\boldsymbol{\Delta\theta}: \boldsymbol{s+\Delta\theta} \in \boldsymbol{N(\theta)}} \sum_{\boldsymbol{t=0}} \boldsymbol{\gamma^t} \boldsymbol{R}(s + \boldsymbol{\Delta\theta}) \left(\boldsymbol{P_i^{t+1}}\right)_{\boldsymbol{\theta_n}, \boldsymbol{\theta_n} + \boldsymbol{\Delta\theta}} \,. \tag{3}$$

Thus, $\boldsymbol{X_t} = \boldsymbol{s} + \boldsymbol{\Delta\theta_t}$ is a random walk with initial state $s$ and transition probability matrix $\boldsymbol{P_i}$. Note, that the previously found optimal state $\boldsymbol{\theta_n}$ (in the former microscopic episode) is not the initial state, but single point at which the ground truth (true value of the return function $\boldsymbol{R(\theta_n)}$) is known exactly.

After the value function has been evaluated, the optimal new parameter is

$$\boldsymbol{\theta_{n+1}} = \underset{\mathbf{s} \in \mathbf{N}(\boldsymbol{\theta_n})}{\mathbf{argmin}} \boldsymbol{V}(\mathbf{s}). \quad (4)$$

After changing to the new optimal parameter $\theta_{n+1}$, the policy may be changed. So even though during the microscopic episodes, the policy stays fixed, this is an actor-critic method, in the sense that any chosen $\boldsymbol{\pi_i}$ is replaced until the value function is minimized by the actor's optimal choice of the policy.

**The surrogate model**. Since the reward at parameter $\boldsymbol{\theta_n} + \boldsymbol{\Delta\theta_t}$ is not known exactly for $\boldsymbol{\Delta\theta_t} \neq \mathbf{0}$ (since the exact flow-field $\boldsymbol{u}(\boldsymbol{\theta_n} + \boldsymbol{\Delta\theta_t})$ is not known), it must be estimated. Note that $\boldsymbol{R}$ and therefore $\boldsymbol{V}$'s dependence on $\boldsymbol{\theta_n}$ is via $\boldsymbol{u}(\boldsymbol{\theta_n})$. The reward function $\boldsymbol{R}(\boldsymbol{\theta_n})$ has expansion

$$\boldsymbol{R}(\boldsymbol{\theta_n} + \boldsymbol{\Delta\theta_n}) = \boldsymbol{R}(\boldsymbol{\theta_n}) + \boldsymbol{\Delta\theta_n} \cdot \nabla_{\boldsymbol{\theta_n}} \boldsymbol{R}(\boldsymbol{\theta_n}) + \boldsymbol{O}\left(\boldsymbol{\Delta\theta_n^2}\right),$$

which could be used to define an *approximation* $\widehat{\boldsymbol{R}}(\boldsymbol{\theta_{n+1}})$ if a physical surrogate for the gradient of the flow field is available. Not being in possession of such a surrogate model, we turn to an interpolation model defined from a set of ground truth simulations, however few, but increasing in number as the mesoscopic path increases, with every newly found optimal parameter $\boldsymbol{\theta_n}$, where another simulation is carried out.

Namely, in addition to knowing the exact value of $\boldsymbol{R}$ at $\boldsymbol{\theta_n}$, we pick a few extra points $\boldsymbol{s_j}$ in the neighborhood $\boldsymbol{N}(\boldsymbol{\theta_n})$ in addition to the middle point $\boldsymbol{\theta_n}$ of the bounding four lines to obtain the ground truth values $\boldsymbol{R}(\boldsymbol{s_j})$. These are used in an interpolation model for the estimate $\widehat{\boldsymbol{R}}_{\boldsymbol{\theta_n}}(\boldsymbol{s})$ (see Figure 5).

**Markov Chain Monte Carlo**. Finally, to approximate the expected values with respect to the transition matrices $\boldsymbol{P_i}$ belonging to the policy $\boldsymbol{\pi_i}$, we use the MCMC-sampling technique given by the Metropolis algorithm [25]. We use the transition probabilities, which, together with the initial node define the measure of the expected value functional in (3), by

$$(\boldsymbol{P_i})_{\boldsymbol{\theta},\boldsymbol{\theta'}} = \frac{\mathbf{1}_{A_i}(\boldsymbol{\Delta\theta'})\, e^{-\beta\left(\mathrm{V}(\boldsymbol{\theta'})-\mathrm{V}(\boldsymbol{\theta})\right) \vee \mathbf{0}}}{\sum_{\boldsymbol{\theta''}} \mathbf{1}_{A_i}(\boldsymbol{\theta''}-\boldsymbol{\theta})\, e^{-\beta\left(\mathrm{V}(\boldsymbol{\theta''})-\mathrm{V}(\boldsymbol{\theta})\right) \vee \mathbf{0}}}, \quad (5)$$

with the sum over $\theta''$ ranging over all elements of $\boldsymbol{N}(\boldsymbol{\theta})$. Note $(\boldsymbol{P_i})_{\boldsymbol{\theta},\boldsymbol{\theta'}}$ depends on the value function, and $\boldsymbol{V}$'s definition () depends on $\boldsymbol{P}$. This sets the scene for a fixed-point iteration to meet the condition of this self-consistency. Just as in simulated annealing, where the minimum of an energy-function is found with the Metropolis algorithm, the extremum of $\widehat{\boldsymbol{V}}_{\boldsymbol{\pi_i}}(\cdot)$ is found here on a suitably chosen neighbourhood $\boldsymbol{N}(\boldsymbol{\theta_n})$ of $\boldsymbol{\theta_n}$. If $\boldsymbol{V}_{\boldsymbol{\pi_i}}$ can be estimated sufficiently well on this $\boldsymbol{N}(\boldsymbol{\theta_n})$, then the estimate of $\boldsymbol{\theta}^*_{\boldsymbol{n+1}}$ will be a sufficiently good initial value in the subsequent cycle.

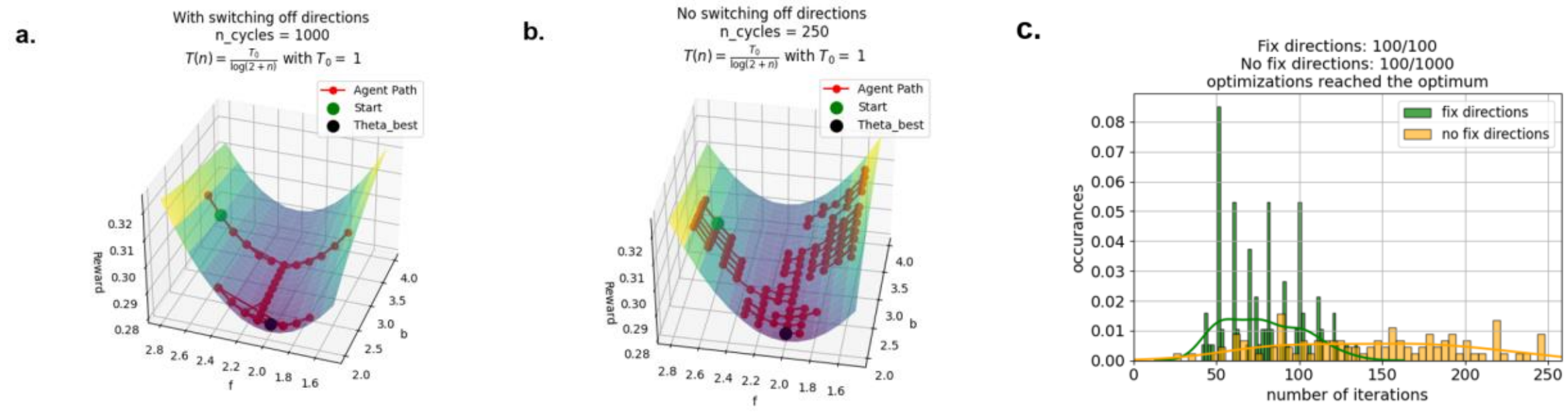

*Figure 4: a.) Showing a sampled Metropolis random walk with logarithmic cooling schedule under the restriction of **fixed** most significant dimension (first $f$, then b). b.) Showing a sampled 2-dimensional (**no fix**) Metropolis random walk with the same cooling schedule. c.) Showing the associated numbers of steps reaching the global minimum the first time.*

It is seen that the mean number of steps to reach the global minimum is much higher in the case of the Metropolis random walk with fixed dimension. The reduction in computational complexity will be used advantageously in the estimation of the value function.

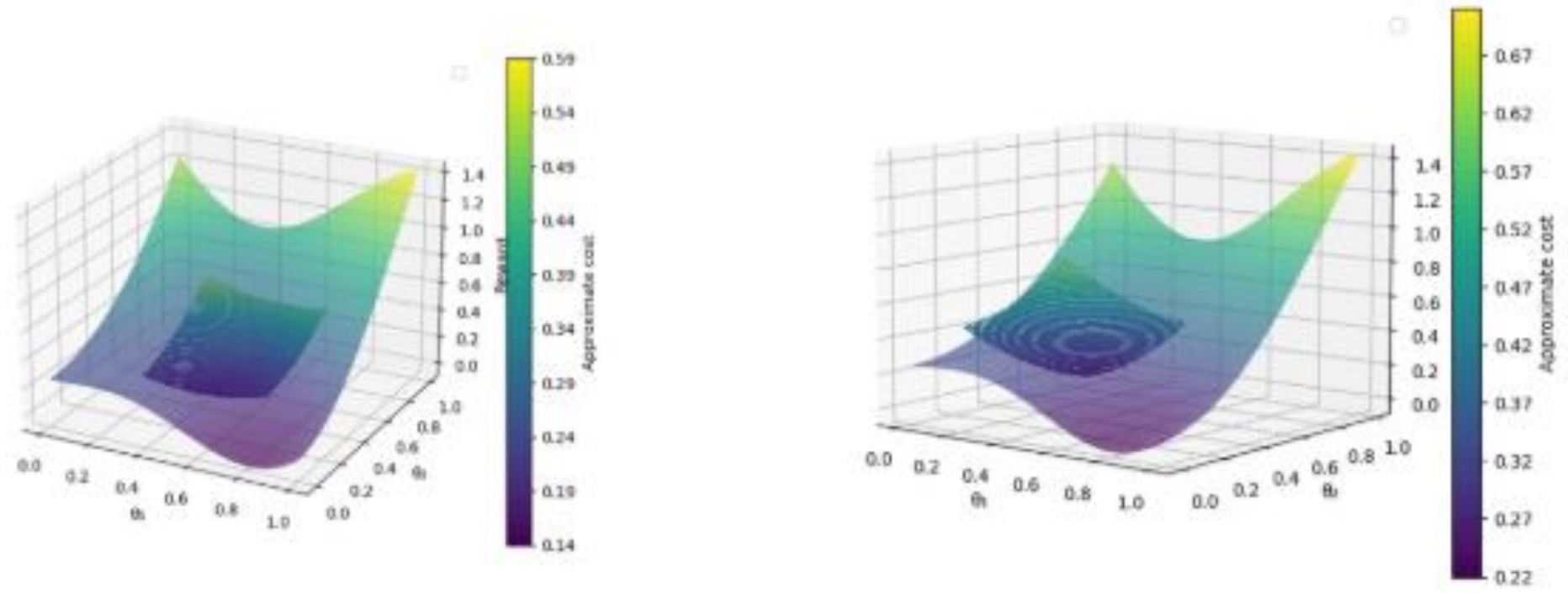

*Figure 5: Example of a possible interpolation surrogate model depending on few (k) ground-truth value of $\boldsymbol{R}(\boldsymbol{s_j})$, with the set chosen $\boldsymbol{s_j}$ from the neighbourhood $\boldsymbol{N}(\boldsymbol{\theta})$: $\widehat{\boldsymbol{R}}_{\boldsymbol{\theta}}(\boldsymbol{s} = (\boldsymbol{x}, \boldsymbol{y})) = \boldsymbol{a}(\boldsymbol{x} - \boldsymbol{\theta_x})^2 + \boldsymbol{b}(\boldsymbol{y} - \boldsymbol{\theta_y})^2 + \boldsymbol{c}(\boldsymbol{x} - \boldsymbol{\theta_x})(\boldsymbol{y} - \boldsymbol{\theta_y}) + \boldsymbol{d}(\boldsymbol{x} - \boldsymbol{\theta_x}) + \boldsymbol{e}(\boldsymbol{y} - \boldsymbol{\theta_y})$. In the experiment carried out in the last section, $k = 4$, and the points $\boldsymbol{s_j}$ have been chosen to be the corner points of the current rectangular region.*

**Policy evaluation.** Note that during a microscopic episode, the policy and therefore the action set $\boldsymbol{A_i}$ stays the same. At the end of each such episode, the following definition is used to decide whether a switch to another $i$ should be carried out.

**Definition 1**: Given any $\boldsymbol{\varepsilon} > \boldsymbol{0}$, a root of the polynomial $\boldsymbol{p}(\boldsymbol{x})$ in the reduced PARSEC representation of an airfoil is called $\boldsymbol{\varepsilon}$-*stable* for the parameter vector $\boldsymbol{\theta}$, if

$$\left|\boldsymbol{\partial}_{\boldsymbol{\theta_i}} \boldsymbol{R}(\boldsymbol{\theta})\right| < \boldsymbol{\varepsilon}. \quad (6)$$

We use this in a relative way to the changes within the neighbourhood. Namely, it holds that the maximum of the $i$'th gradient in $\boldsymbol{N}(\boldsymbol{\theta_n})$ ($i$'th direction) is less than the $\boldsymbol{\epsilon}$'th part of the maximal gradient in $\boldsymbol{N}(\boldsymbol{\theta_n})$ then the $i$'th direction is 'switched off' by passing to the corresponding reduced parameter action set. Since the gradient isn't known exactly (just like the objective function), it must be estimated. See the definition of the algorithm for the precise definition in this case.

**Estimation of the Value function**. Like in the temporal difference learning techniques of RL leading to Bellman's equation [17], our Monte Carlo approach needs a solution of Equation (3) defining the value function $\boldsymbol{V}(\mathbf{s})$, which must also fulfil equation (5). The solution can be approximated by fixed point iterations using these definitions. Using $j$ as the index of the $j$'th iteration,

$$\boldsymbol{V^{(j+1)}(s) = E[\sum_{t=0}^{\infty} \gamma^t R(s + \Delta\theta_t^{(j)})]}, \tag{8}$$

where $\boldsymbol{X_t = s + \Delta\theta_t^{(j)}}$ is the random walk defined by the initial distribution $\boldsymbol{P[X_0 = s] = 1}$ and transition probability $\boldsymbol{P_{\theta,\theta'}^{(j)} = \left(P_i^{(j)}\right)_{\theta,\theta'}}$ for some $i$ (fixed during the microscopic episode) given by

$$\boldsymbol{\left(P_i^{(j)}\right)_{\theta,\theta'} = \frac{1_{A_i}(\Delta\theta')\, e^{-\beta_j \left(V^{(j)}(\theta') - V^{(j)}(\theta)\right) \vee 0}}{\sum_{\theta''} 1_{A_i}(\theta'' - \theta)\, e^{-\beta_j \left(V^{(j)}(\theta'') - V^{(j)}(\theta)\right) \vee 0}}}, \tag{9}$$

where $\boldsymbol{\beta_j}$ is some decreasing function, e.g. $\boldsymbol{\beta_j = \frac{1}{T_0 \cdot log(2+j)}}$ , representing the cooling schedule of the stochastic annealing method [25]. (Index $i$ represents the policy, and $j$ the number of the iteration.)

Like the value iteration used for solving Bellman's equation (but without *conditioning* on a specific action ([17], Chap. 3.6), a fixed-point iteration may be used to estimate the expression given in (3). If multiplied (and divided) by $1-\gamma$, it is a specific (non-arithmetic) mean over values from the return function (divided by $1-\gamma$), taken at arguments (parameters) visited by the agent over (future) time. Since our return function is defined on a bounded set and therefore bounded from above and below, the sequence of fixed-point iterations remains within these bounds. Non-convergence therefore only happens if there are multiple converging sub-sequences [26]. Figure 4 shows a one-dimensional computation with a fictitious reward function given by $R(x) = (x+2)^2(x+1)^2(x-1)^2$ with 30 iterations. It is seen that the result converges towards a function with more pronounced V-shaped local (pointed) extrema and sharp dividing points between the basins belonging to each local minimum (see Figure 6). Note that again, we consider small values of the function $V^{(j)}(x)$ as favourable – so that the 'reward' is a penalty, and $R(x)$ is from now on called the **objective function**.

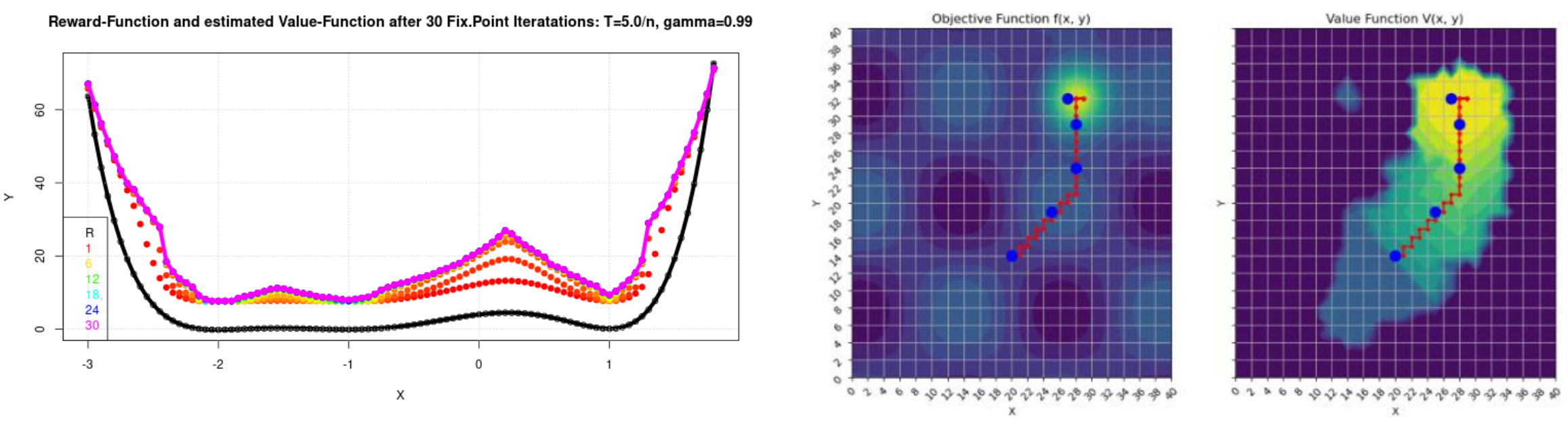

*Figure 6: Left – Computation of 1-dimensional fixed-point iterations of the sequence $V^{(i)}(s)$ with $V^{(0)}(s) = R(s)$. It is seen that flat wells of $R(s)$ turn into stronger V-shaped minima, while saddle points lose their round shape to a peaked form. Right – computation of an estimate of the limiting function of the fixed-point iterations of (3) and (5) of the return function shown in Figure 3, on the right.*

**Optimal Strategy.** Having obtained an estimate of the value-function, we can use the Bellman principle to derive a sequence of optimal parameter-changes which can be interpreted in the physical sense to characterize the importance of the specific geometric features (PARSEC parameters) in finding the optimal aerodynamic shape. From the sequence of policy evaluations, it will become apparent which parameters are worth including in the local optimizations during the microscopic episodes. This will be specific to the different regions in parameter space. Learning which remaining parameters are locally important for the significant part of the parameter adaptation ($\boldsymbol{\theta_n \rightarrow \theta_{n+1}}$) is valuable from a physical perspective: It indicates the significance of the geometric feature controlled by these remaining parameters on the effects expressed by the value function.

**Pseudo-Code**: The protocol of our method reads as follows:

---

**Dimension Reduction**

---

1. **Initialisation:** Start with an initial value for the parameter vector: $\boldsymbol{\theta}_0$, set $\boldsymbol{n} = \mathbf{0}$.
   Choose dimensions of neighbourhood $\boldsymbol{N}(\boldsymbol{\theta}_0)$, a discount factor $\boldsymbol{\gamma}$, and a cooling schedule $\boldsymbol{\beta}_j = \frac{\mathbf{1}}{\boldsymbol{T}(j)}$.
   Choose the action set $\boldsymbol{A}_i$ which allows all coordinates of the parameter vector to change.

2. **Surrogate model:** Choose an estimator $\boldsymbol{s} \mapsto \widehat{\boldsymbol{R}}_{\boldsymbol{\theta}}(\boldsymbol{s})$ for the objective function $\boldsymbol{R}(\boldsymbol{s})$, which fulfils
   $$\widehat{\boldsymbol{R}}_{\boldsymbol{\theta}_n}(\boldsymbol{\theta}_n) = \boldsymbol{R}(\boldsymbol{\theta}_n).$$

3. **Value function estimation**: For all $\boldsymbol{s} \in \boldsymbol{N}_\rho(\boldsymbol{\theta}_n)$, estimate the value function $\boldsymbol{V}(\boldsymbol{s})$ using (9),
   $$\widehat{\boldsymbol{V}}^{(j+1)}(\boldsymbol{s}) = \boldsymbol{E}\left[\sum \boldsymbol{\gamma}^t\, \widehat{\boldsymbol{R}}\left(\boldsymbol{s} + \Delta\boldsymbol{\theta}_t^{(j)}\right) \middle| \Delta\boldsymbol{\theta}_0 = \mathbf{0}\right] = \sum_{t=0}^{\infty} \boldsymbol{\gamma}^t \sum_{\theta' \in N_\rho(\theta_n)} \widehat{\boldsymbol{R}}(\boldsymbol{\theta}') \left(\boldsymbol{P}_i^{(j)}\right)^t_{\theta,\theta'},$$
   iteratively over $\boldsymbol{j} \in \boldsymbol{Z}_+$, where $(\boldsymbol{P}_i^{(j)})$ is given by (3), and $\widehat{\boldsymbol{V}}^{(\mathbf{0})}(\boldsymbol{s}) = \widehat{\boldsymbol{R}}(\boldsymbol{s})$.

4. **Microscopic optimum**: With the estimated value function, define the optimum in the current microscopic episode
   $$\boldsymbol{\theta}_{n+1} = \underset{\mathbf{s} \in \boldsymbol{N}(\boldsymbol{\theta}_n)}{\mathbf{argmin}}\, \widehat{\boldsymbol{V}}(\mathbf{s}).$$

5. **Policy evaluation**: Decide which degrees of freedom to allow for changing in the next episode.
   Use **$\epsilon$-stability**: For every dimension $j$, let $\boldsymbol{M}_j = \{\boldsymbol{\theta} \in \boldsymbol{N}(\boldsymbol{\theta}_{n+1}) |\, \boldsymbol{\theta} = \boldsymbol{\theta}_{n+1} + t \cdot \boldsymbol{e}_j, t \in (-\infty, \infty)\}$.
   If
   $$\max_{s \in M_j}\left(\widehat{\boldsymbol{R}}_{\theta_{n+1}}(\boldsymbol{\theta}_{n+1}) - \widehat{\boldsymbol{R}}_{\theta_{n+1}}(\boldsymbol{s})\right) < \epsilon \cdot \max_{s \in N(\theta_{n+1})}\left(\widehat{\boldsymbol{R}}_{\theta_{n+1}}(\boldsymbol{\theta}_{n+1}) - \widehat{\boldsymbol{R}}_{\theta_{n+1}}(\boldsymbol{s})\right),$$
   then remove this dimension from the action set. Increase the size of $\boldsymbol{N}(\boldsymbol{\theta}_{n+1})$ in the remaining dimensions, such that the size of the resulting neighbourhood approximately matches that of $\boldsymbol{N}(\boldsymbol{\theta}_n)$.

6. Increase $\boldsymbol{n}$ by one. Goto 3. Repeat until $\boldsymbol{\theta}_{n+1} \cong \boldsymbol{\theta}_n$, according to some accuracy criterion.

---

## 3. Use Case

For an airfoil in a channel with vertical periodic boundary conditions, given inclined inflow and constant pressure outflow conditions of a fluid described by Stoke's equation, we first tested the Reduced PARSEC representation with a symmetric profile and an additively superimposed function defining the camber line. Namely, $\boldsymbol{Z}_{up/lo}(\boldsymbol{x}) = \sqrt{\boldsymbol{x}} \cdot \boldsymbol{p}_{up/lo}(\boldsymbol{x}) + \boldsymbol{CAM}(\boldsymbol{x})$ with $\boldsymbol{x} \in [\mathbf{0}, \mathbf{1}]$, assuming a chord length equal to one, where

$\boldsymbol{p}(\boldsymbol{x}) = \boldsymbol{c}(\boldsymbol{x} - \mathbf{1})(\boldsymbol{x} - \boldsymbol{b})$ with $\boldsymbol{b} > \mathbf{1}$, and $\boldsymbol{CAM}(\boldsymbol{x}) = \boldsymbol{e} \cdot \boldsymbol{x}(\boldsymbol{f} - \boldsymbol{x})$ width $\boldsymbol{f} \geq \mathbf{1}$.

The parameters other than $\boldsymbol{f}, \boldsymbol{b}$ were chosen in the following way: $\boldsymbol{c}_{up} = \frac{\mathbf{1}}{\mathbf{2}\boldsymbol{b}^2}$, $\boldsymbol{c}_{lo} = \frac{\boldsymbol{b}}{\mathbf{2}}$, $\boldsymbol{e} = \mathbf{0.3}$. The specific dependence on $\boldsymbol{b}$ makes the width of the profile stay approximately the same.

Figure 3 (Right) and Figure 4 show typical results of the exactly calculated objective function $\boldsymbol{R}(\boldsymbol{\theta})$ with an extremum appearing at the boundary of the constructable range of parameter-pairs in the $\boldsymbol{f} \times \boldsymbol{e}$ plane.

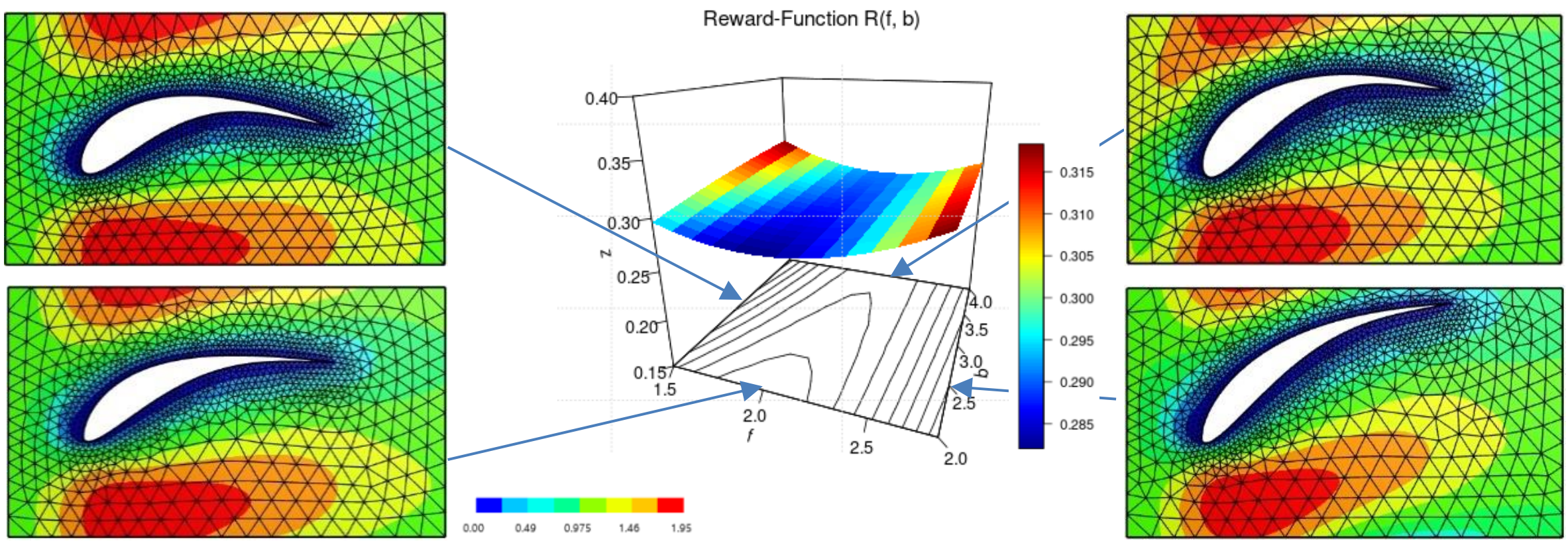

*Figure 7: Cost function* $R(\theta) = \frac{1}{|E|}\sum_{x\in E}\frac{|u_2(x)|^2}{\sqrt{|u_1(x)|^2+|u_2(x)|^2}} + \max_{x\in E}|u(x)| - \min_{x\in E}|u(x)|$ *with* $\theta = (f, b) \in [1.5, 2.8] \times [2.0, 4.0]$ *and norms of the velocity flow-field (in the surrounding graphs). At about* $(f, b) = (2.0, 2.0)$ *there is a local minimum due to the air foil's thickness cannot be reduced further (*$b \geq 2$*). The inward flow from the left in the flow-field simulations has an upward component direction:* $u_{in} = \left(1.0\frac{m}{s}, 0.75\frac{m}{s}\right)$*. The unit of the two scales (flow-fields, Reward function) is both* $\frac{m}{s}$*.*

Figure 7 shows the evaluation of the objective function for plausibility of this (toy) model. The camber parameter $f$ controls the vertical position of the trailing edge, while the form parameter $b$ controls the curvedness of the air foil.

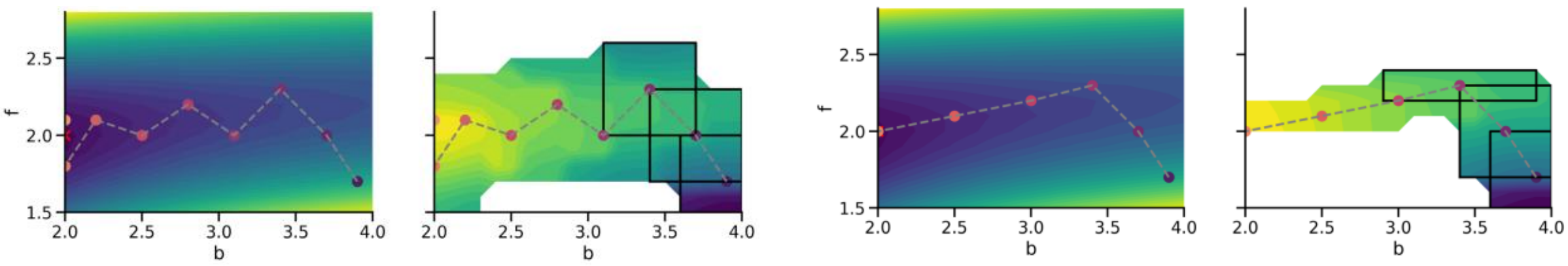

*Figure 8: Left* - Mesoscopic path of length 8 with objective function (left) and estimated value function on patches of 7x7 grid-points (right). Right – Mesoscopic path of length 5 with objective function (left) and estimated value function on sequence of patches (at first) of size 3x7 (2 times) and afterwards of size 7x3 (3 times). The mesoscopic paths are induced by the local minima of the value function estimations on the neighbourhoods $\boldsymbol{N(\theta_n)}$**.** The adaptive passing to rectangles aligned with the valley-shape of the objective function accelerates the detection of the minimum.

**Experiment 1**: We now present calculations of optimization runs in this two-dimensional parameter space with various starting points $\theta_0 = (f = f_0, b = b_0)$ and neighbourhood-sizes. We vary the size of the neighbourhoods $N(\theta_n)$. In all cases we compare the discrete time steps necessary to reach the minimum with and without fixing of the parameters. In other words, we compare allowing the adaption of the policy to switching from $A_3$ to the reduced action sets $A_1, A_2$ with the version in which only the Actions from $A_3$ are at the agent's disposal.

| Run | Starting Point | Quadratic Dimensions $\boldsymbol{N(\theta)}$ | Length of Mesoscopic Path | Rectangular Dimensions $\boldsymbol{N(\theta)}$ | Length of Mesoscopic Path |
|---|---|---|---|---|---|
| 1 | (2.2, 1.7) | (7x7) | 1 | (3x3) | 1 |
| 2 | (2.2, 2.6) | $(7x7)^3$ | 3 | $(7x7)^2$, $(11x1)^2$ | 4 |
| 3 | (3.0, 1.7) | $(7x7)^5$ | 5 | $(7x7)^2$, (11x1) | 3 |
| 4 | (3.0, 2.6) | $(7x7)^5$ | 5 | $(7x7)^2$, $(11x1)^3$ | 5 |
| 5 | (3.9, 1.7) | $(7x7)^8$ | 8 | $(7x7)^2$, $(11x1)^3$ | 5 |
| 6 | (3.9, 2.6) | $(7x6)^7$ | 7 | $(7x7)^2$, $(11x1)^3$ | 5 |

**Experiment 2**: From the same (remote) starting point $\theta_0 = (f_0, b_0) = (9.7, 3.9)$ the **length of the mesoscopic path** as well as the **number of iterations** needed for a converged value function estimation is compared for the case of a quadratic (2-dimensional) $\boldsymbol{N(\theta_n)}$ with a sequence of width-one rectangles (1-dimensional) aligning with the surrogate $\boldsymbol{\hat{R}}_{\theta_n}$ according to the Dimension-Reduction algorithm defined in Section 2.

| Radius | Dimensions of quadratic $N(\theta_n)$ | Mesoscopic Path length fixed dimension | # Iterations needed for estimating $V_{\theta_n}(s)$ | Resize | Dimensions of rectangular $N(\theta_n)$ | Mescopic Path length variable dimension | # Iterations needed for estimating $V_{\theta_n}(s)$ |
|---|---|---|---|---|---|---|---|
| 1 | 3x3 | 25 | 48 | | 3x1 -> 1x3 | 6 | 10 |
| 2 | 5x5 | 12 | 26 | | 5x1 -> 1x5 | 14 | 27 |
| 3 | 7x7 | 9 | 24 | | 7x1 -> 1x7 | 11 | 23 |
| 4 | 9x9 | 10 | 27 | | 9x1 -> 1x9 | 9 | 19 |
| 5 | 11x11 | 7 | 18 | | 11x1 -> 1x11 | 7 | 14 |

The resize in Experiment 2 refers to the second part of Step 5 of the Dimension Reduction Algorithm: The arrow heads indicate how much larger the rectangle must be increased on the long side to achieve minimal improvement in performance (either in number of time steps or number of iterations to approximate $V(s)$).

**Observations:**

- From Experiment 1, runs 3 to 6, it is seen that it is typically advantageous for the reduction of the mesoscopic path length to use the adaption of the neighbourhood.
- From Experiment 2, it is seen that after reduction of the $\epsilon$-stable parameters, with an increase of the remaining degrees of freedom, as indicated by the arrowheads, an improvement of mesoscopic path length as well as number of approximation steps of the value function is achieved.

## 4. Discussion and Conclusion

We have defined a reinforcement learning assisted optimization strategy to reduce the computational complexity of the numerical calculations of aerodynamic shape optimization. The method uses dimensionality reduction and a sequence of local approximations of the objective function associated with the quality of different aerodynamic shapes. The simple model problem discussed in the experimental section reflects the improvements which can be expected from adapting the choice of neighbourhood around intermediate results of the local optima. The key advantages of this method are:

- The value function estimation from RL replaces the estimated objective function with a more useful quantity to be minimized: The resulting shapes of these value functions increase the distinction of the local minima within each element of the sequence of neighbourhoods improving their estimation (i.e., the local minima on the neighbourhoods $N(\theta_n)$ are closer to the true minima of the real reward function $R(\theta)$ on these regions).
- The possibility to adapt the sizes of the neighbourhoods $N(\theta_n)$ by policy evaluation at the end of each microscopic episode makes the proposed method an adaptive optimization method capable of reducing the calculation load to the necessary minimum in terms of considered degrees of freedom.
- It is seen from the experiment, that as an intermediate step towards complete dimensionality reduction, a speed-up is already visible if the rectangular regions $N(\theta_n)$ are changed towards aligning with the local structure of the objective function $R(\theta)$.

Our outlook is to use the proposed scheme to solve 3-dimensional aerodynamic shape optimization problems for the design and increase in efficiency of air source heat pumps [1].

## 5. Acknowledgement

The main contributor to this research is the FFG (Austrian Applied research fund) project “**AIrFoil”** (FFG-No. 915010). The research reported in this paper has also been partly funded by the Federal Ministry for Innovation, Mobility and Infrastructure (BMIMI), the Federal Ministry for Economy, Energy and Tourism (BMWET), and the State of Upper Austria in the frame of the SCCH competence center INTEGRATE [(FFG grant no. 892418)] in the COMET - Competence Centers for Excellent Technologies Programme managed by Austrian Research Promotion Agency FFG.